\documentclass[journal=jacsat,manuscript=article]{achemso}

\usepackage[version=3]{mhchem} % Formula subscripts using \ce{}
\usepackage{siunitx}
\usepackage{graphicx}
\usepackage{bbm}
\usepackage{amsfonts}
\usepackage{xcolor}
\usepackage{lineno}
\usepackage{etoolbox}
\usepackage{titlesec}
\usepackage{hyperref}       % hyperlinks
\usepackage{url}            % simple URL typesetting
\usepackage{booktabs}       % professional-quality tables
\usepackage{nicefrac}       % compact symbols for 1/2, etc.
\usepackage{microtype}      % microtypography
\usepackage{amsmath}
\usepackage{subcaption}
\usepackage{wrapfig}
\usepackage{enumitem}
\usepackage{multirow}
\usepackage{makecell}

\newcommand{\bftab}{\fontseries{b}\selectfont}

\DeclareUnicodeCharacter{2212}{-}

\author{Yuyang Wang}
\affiliation[meche]
{Department of Mechanical Engineering, Carnegie Mellon University, Pittsburgh, PA, USA}
\author{Rishikesh Magar}
\affiliation[meche]
{Department of Mechanical Engineering, Carnegie Mellon University, Pittsburgh, PA, USA}
\author{Chen Liang}
\affiliation[cheme]
{Department of Chemical Engineering, Carnegie Mellon University, Pittsburgh, PA, USA}
\author{Amir Barati Farimani}
\email{barati@cmu.edu}
\affiliation[meche]
{Department of Mechanical Engineering, Carnegie Mellon University, Pittsburgh, PA, USA}
\alsoaffiliation[cheme]
{Department of Chemical Engineering, Carnegie Mellon University, Pittsburgh, PA, USA}

\title[iMolCLR]{Improving Molecular Contrastive Learning via Faulty Negative Mitigation and Decomposed Fragment Contrast}

\abbreviations{IR,NMR,UV}

\begin{document}

\begin{abstract}

Deep learning has been a prevalence in computational chemistry and widely implemented in molecule property predictions. Recently, self-supervised learning (SSL), especially contrastive learning (CL), gathers growing attention for the potential to learn molecular representations that generalize to the gigantic chemical space. Unlike supervised learning, SSL can directly leverage large unlabeled data, which greatly reduces the effort to acquire molecular property labels through costly and time-consuming simulations or experiments. However, most molecular SSL methods borrow the insights from the machine learning community but neglect the unique cheminformatics (e.g., molecular fingerprints) and multi-level graphical structures (e.g., functional groups) of molecules. In this work, we propose iMolCLR: \textbf{i}mprovement of \textbf{Mol}ecular \textbf{C}ontrastive \textbf{L}earning of \textbf{R}epresentations with graph neural networks (GNNs) in two aspects, (1) mitigating faulty negative contrastive instances via considering cheminformatics similarities between molecule pairs; (2) fragment-level contrasting between intra- and inter-molecule substructures decomposed from molecules. Experiments have shown that the proposed strategies significantly improve the performance of GNN models on various challenging molecular property predictions. In comparison to the previous CL framework, iMolCLR demonstrates an averaged 1.3\% improvement of ROC-AUC on 7 classification benchmarks and an averaged 4.8\% decrease of the error on 5 regression benchmarks. On most benchmarks, the generic GNN pre-trained by iMolCLR rivals or even surpasses supervised learning models with sophisticated architecture designs and engineered features. Further investigations demonstrate that representations learned through iMolCLR intrinsically embed scaffolds and functional groups that can reason molecule similarities.

\end{abstract}

%%%%%%%%%%%%%%%%%%%%%%%%%%%%%%%%%%%%%%%%%%%%%%%%%%%%%%%%%%%%%%%%%%%%%

\section{Introduction}

Recent years have witnessed the development of computational molecule design and property prediction driven by deep learning (DL) \cite{lecun2015deep, schmidhuber2015deep}, owing to the ability to perform fast and accurate computation \cite{duvenaud2015convfingerprint, butler2018machine, wang2021efficient, alquraishi2021differentiable}. Several works build deep neural networks on top of cheminformatics fingerprints to predict molecular properties \cite{unterthiner2014deep, ma2015deep, vamathevan2019applications}. DL methods have been also implemented on string-based molecular embeddings, like SMILES \cite{weininger1988smiles} and SELFIES \cite{krenn2020self}, for molecule design \cite{xu2017seq2seq, gomez2018automatic}. However, both fingerprints and string embeddings can neglect important structural information of molecules. Recently, graph neural networks (GNNs) \cite{kipf2016semi, xu2018how} are developed to learn representations from non-Euclidean graphs of chemical structures, where each node in molecule graphs are defined as an atom and each edge represent chemical bond or adjacency of atoms \cite{jiang2021could}. Modern GNNs rely on message-passing to aggregate neighboring node information within the graph and have been introduced to predict various properties from molecule graphs \cite{gilmer2017neural, yang2019analyzing}. Aggregation based upon a continuous filter is also developed for GNNs to model quantum interactions within molecules \cite{schutt2018schnet, lu2019molecular}. Attention mechanism \cite{vaswani2017attention} has also been leveraged in node aggregation for better prediction accuracy and model interpretability \cite{xiong2019pushing, ying2021do}. Instead of only considering molecules as two-dimensional (2D) graphs, several works have built GNNs in use of 3D molecular conformations with equivariant aggregation \cite{klicpera2020directional, fuchs2020se3, liu2021spherical, jing2021learning}. 

Despite the success of DL in computational chemistry, the potential is greatly limited by the availability of labeled data, as the collection of molecule properties usually requires time-consuming and costly lab experiments or simulations \cite{wu2018moleculenet}. Moreover, it is challenging for DL models trained on such limited data to generalize among the gigantic chemical space \cite{oprea2001chemography}, which significantly restricts real-world applications like drug discovery and material design. To address this, self-supervised learning (SSL) \cite{hadsell2006dimensionality,doersch2017multi} have been investigated to utilize the large unlabelled molecule data and learn representations generalizable to various down-stream applications. Motivated by the success of SSL in language models, transformer-based models \cite{vaswani2017attention}, like BERT \cite{devlin2018bert}, have been implemented to learn representations from large SMILES database \cite{wang2019smiles,chithrananda2020chemberta,fabian2020molecular,flam2021keeping,ross2021large}. Apart from language models, SSL has also been developed for representation learning from molecule graphs. Liu et al. \cite{Liu2019NGramGS} embed molecules to N-gram representations by assembling the vertex embedding in short walks. Hu et al. \cite{Hu2020Strategies} propose both node-level and graph-level GNN pre-training strategies. The former includes self-supervised context prediction and attribute masking, while the latter is based on supervised property prediction, which is still limited by label availability. Following the insights, Rong et al. \cite{rong2020self} propose contextual property and graph-level motif predictions as SSL tasks combined with a transformer-based model. Besides node-level pre-training, Zhang et al. \cite{zhang2021motif} introduce a motif-level SSL strategy, which builds motif trees from molecules and performs motif generative pre-training. Additionally, contrastive learning (CL) \cite{he2020momentum, chen2020simple, caron2020unsupervised, chen2021exploring, zbontar2021barlow}, which learns representation through contrasting positive pairs against negative pairs, has been a prevalence in representation learning \cite{bengio2021deep} and implemented to graphical data \cite{you2020graph}. Wang et al. \cite{wang2021molclr} propose MolCLR, a CL framework for molecular representation learning, as well as three augmentation strategies to generate contrastive pairs. Zhang et al. \cite{zhang2020motif} further leverage frequently-occurring subgraph patterns and perform CL on subgraph level. Liu et al. \cite{liu2021pre} and St{\"a}rk et al. \cite{stark20213d} perform contrastive training between 2D topological structures and 3D geometric views to learn molecular representations with 3D information embedded. Also, Zhu et al. \cite{zhu2021dual} develop multi-view CL between SMILES strings and molecule graphs, encoded by transformer and GNN, respectively. Although CL has demonstrated the effectiveness in molecular representation learning, these methods assume all the other molecules are equal negative pairs in contrast with a given anchor, which introduces faulty negatives. Faulty negatives are instances that are supposed to be similar with the anchor while considered as negative instances in CL \cite{morgado2021robust}. 
Such faulty negative instances harm the robustness and performance of CL pre-trained model on downstream property prediction tasks. Additionally, the previous motif-level CL learns a motif dictionary and trains a sampler to sample subgraphs within each molecule \cite{zhang2020motif}, which may ignore unique chemical substructure patterns. Chemical substructures of molecules contain functional groups that are critical to various molecular properties, which can provide multi-level information for representation learning. 

\begin{figure}[t!]
    \centering
    \includegraphics[width=\textwidth, keepaspectratio=true]{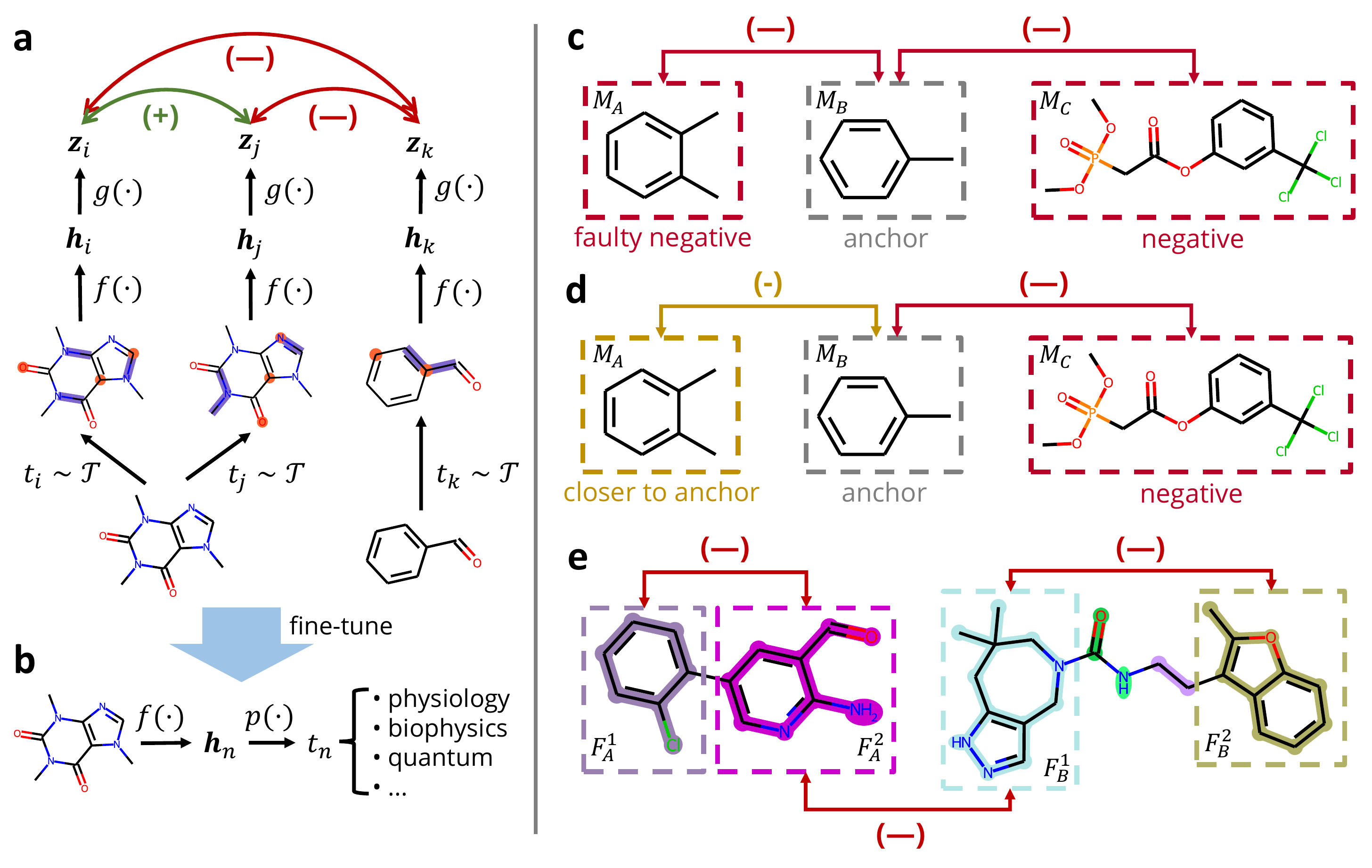}
    \caption{Overview of the proposed iMolCLR framework. (a) CL pre-training pipeline for molecular representation learning. (b) Fine-tuning of pre-trained model for various property predictions. (c) An example of faulty negative introduced by ordinary CL methods due to treating all negative instances equivalently. (d) In iMolCLR, latent vectors of molecules are repelled based upon the similarities between different molecule pairs to mitigate faulty negatives. (e) Fragment-level contrast on decomposed molecular substructures in iMolCLR. }
    \label{fig:overview}
\end{figure}

In this work, we propose iMolCLR: improvement of molecular contrastive learning through mitigating faulty negative instances and contrasting decomposed chemical fragments as shown in Figure~\ref{fig:overview}. A GNN encoder is first trained on large unlabeled data to learn expressive molecular representations via contrasting positive pairs against negative pairs in a self-supervised manner (Figure~\ref{fig:overview}a). The pre-trained GNN is fine-tuned on downstream datasets to predict a wide variety of molecular properties (Figure~\ref{fig:overview}b). Unlike ordinary CL framework that introduces faulty negatives (Figure~\ref{fig:overview}c), iMolCLR does not treat all negative pairs equivalently. On the contrary, similar molecules are encouraged to have closer representations than dissimilar ones (Figure~\ref{fig:overview}d). Apart from molecule-level contrast, different substructures decomposed via breaking retrosynthetically interesting chemical substructures (BRICS) \cite{degen2008art} are also considered as contrastive negative pairs (Figure~\ref{fig:overview}e). Such a decomposition strategy maintains major structural features of compounds and molecular representations are forced to distinguish important functional groups within molecules. Experiments show that iMolCLR pre-training significantly improves the performance of GNN models on challenging molecular property prediction benchmarks. Through fine-tuning, iMolCLR rivals and even surpasses strong supervised learning baselines on multiple classification and regression tasks. Additionally, it shows an overall advantage over other SSL baselines. In particular, iMolCLR outperforms the original CL framework by an average 1.3\% improvement of ROC-AUC on 7 classification benchmarks and an average 4.8\% reduction of the error on 5 regression benchmarks. Additional investigations demonstrate that the proposed CL framework effectively learns intrinsic relations between atoms without meticulously engineered input descriptors. The representations learned through iMolCLR also show an advantage to reason molecule similarities in consideration of scaffold and functional groups in comparison to previous SSL methods. 

%%%%%%%%%%%%%%%%%%%%%%%%%%%%%%%%%%%%%%%%%%%%%%%%%%%%%%%%%%%%%%%%%%%%%

\section{Methods}

\subsection{Contrastive Learning Framework}
\label{sec:framework}

CL aims at learning representations through contrasting positive pairs against negative pairs. We develop the CL framework \cite{chen2020simple, wang2021molclr} containing four components: molecule graph augmentation, GNN-based encoder, non-linear projection head, and contrastive loss, as shown in Figure~\ref{fig:overview}a. Given a batch of $N$ molecules $\{m_1, \dots, m_N\}$, each molecule $m_n$ is augmented into two graphs $G_i$ and $G_j$ through augmentations, $t_i$ and $t_j$, sampled from $\mathcal{T}$, where $i=2n-1$ and $j=2n$. We implement $\mathcal{T}$ as random atom masking of 25\% and random bond deletion of 25\% following widely-used graph augmentation strategies \cite{wang2021molclr, magar2021auglichem}. The masked atoms are colored by orange and deleted bonds are colored by dark blue in Figure~\ref{fig:overview}a. Two graphs augmented from the same molecule compose a positive pair while those from different molecules are negative pairs. The GNN-based encoder $f(\cdot)$ takes in an augmented graph $G_i$ and encodes it to the representation $\pmb h_i$, followed by the non-linear projection head $g(\cdot)$ which maps $\pmb h_i$ into a latent vector $\pmb z_i$. Contrastive loss is applied on the $2N$ latent vectors from the projection head to maximize the agreement between positive pair vectors (e.g., $\pmb z_i$ and $\pmb z_j$) while minimizing the agreement between negative ones (e.g., $\pmb z_i$ and $\pmb z_k$, $\pmb z_j$ and $\pmb z_k$). After pre-training, the model $f(\cdot)$ is fine-tuned to predict various molecular properties of interest as shown in Figure~\ref{fig:overview}b. During fine-tuning, only the GNN encoder $f(\cdot)$ is preserved followed by a randomly initialized prediction head $p(\cdot)$ to map the representation to the target property. 

\subsection{Graph Neural Network}
\label{sec:gnn}

A molecule graph $G$ is defined as $G=(V,E)$, where each node $v \in V$ represents an atom and each edge $e_{uv} \in E$ represents a chemical bond between atoms $u$ and $v$ \cite{bronstein2017geometric}. Each node is featurized as $\pmb x_v$ and each edge is featurized as $\pmb \epsilon_{uv}$, which contain unambiguous input vectors to denote each node and edge like atomic number and covalent bond type. Modern GNNs \cite{kipf2016semi} updates the feature of each node layer-wise through iterative combination and aggregation operations. The update rule for the feature of node $v$ at $k$-th graph convolutional layer, $\pmb h_v^{(k)}$, is given in Equation~\ref{eq:aggregate}: 
\begin{equation}
    \pmb{a}_v^{(k)} = \text{AGGREGATE}^{(k)}\Big(\Big\{\pmb h_u^{(k-1)}: u \in \mathcal{N}(v)\Big\}\Big), \; \pmb h_v^{(k)} = \text{COMBINE}^{(k)}\Big(\pmb h_v^{(k-1)}, \pmb a_v^{(k)}\Big), 
    \label{eq:aggregate}
\end{equation}
where $\mathcal{N}(v)$ denotes the set of all the neighbors of node $v$. Aggregation passes the information of neighboring nodes to $v$ and combination updates the aggregated feature. Each $\pmb h_v^{(0)}$ is initialized by the node feature $\pmb{x}_v$. After $K$ layers of node updates, readout operation integrates all the node features within the graph $G$ to a graph-level feature $\pmb h$ as shown in Equation~\ref{eq:readout}:
\begin{equation}
    \pmb h_G = \text{READOUT}\Big(\Big\{\pmb h_u^{(k)}: v \in G\Big\}\Big).
    \label{eq:readout}
\end{equation}

In this work, we develop our GNN encoder based on Graph Isomorphism Network (GIN) \cite{xu2018how}, a widely-used generic model. To consider edge features, we follow Hu et al. \cite{Hu2020Strategies} to extend node aggregation as $\pmb{a}_v^{(k)} = \sum_{u \in \mathcal{N}(v)} \sigma (\pmb h_u^{(k-1)} + \pmb \epsilon_{uv})$ where $\sigma(\cdot)$ is a non-linear activation function. Combination operation is modeled by summation followed by an MLP as $\pmb h_v^{(k)} = \texttt{MLP}(\pmb h_v^{(k-1)} + \pmb{a}_v^{(k)})$. Readout operation is implemented as an average pooling over all nodes to obtain a graph-level representation for each molecule. 

\subsection{Mitigating Faulty Negatives}
\label{sec:nt-xent}

Contrastive loss \cite{oord2018infonce}, like the Normalized Temperature-scaled Cross-Entropy (NT-Xent) loss \cite{chen2020simple}, aims at representation learning through maximizing the agreement between positive pairs while minimizing the agreement between negative pairs. Given $2N$ latent vectors $\{\pmb z_1, \dots, \pmb z_{2N}\}$ from a batch of $N$ molecules, NT-Xent for a positive pair $(\pmb z_i, \pmb z_j)$ is given in Equation~\ref{eq:ntxent}:
\begin{equation}
    \mathcal{L}_{i,j} = - \log \frac{\exp(\cos(\pmb{z}_i, \pmb z_j) / \tau)}{\sum_{k=1}^{2N} \mathbbm{1}_{k \neq i} \exp(\cos(\pmb{z}_i, \pmb z_k) / \tau)},
\label{eq:ntxent}
\end{equation}
where $\tau$ is the temperature parameter and $\cos(\pmb{z}_i, \pmb z_j)=\frac {\langle \pmb{z}_i, \pmb{z}_j \rangle}{\|\pmb{z}_i\| \|\pmb{z}_j\|}$ measures the cosine similarity between two latent vectors. However, such a contrastive loss assumes all negative pairs are equally negative against the anchor $\pmb z_i$, which leads to faulty negatives. Faulty negatives are instances that are similar with the anchor yet are treated as negative instances in contrastive training \cite{morgado2021robust}. An example of faulty negatives introduced in original molecular CL framework is illustrated in Figure~\ref{fig:overview}c. When NT-Xent loss is applied, both the molecule $M_A$ (o-xylene, CID 7237) and molecule $M_C$ (CID 89970782) are trained as equivalent negative instances against the anchor molecule $M_B$ (toluene, CID 1140). However, $M_B$ has much more similar molecular properties to $M_A$ than $M_C$, since $M_A$ and $M_B$ share a similar structure and functional groups. In this case, $M_A$ is a ``faulty negative" as it should not be far away from the anchor in the representation domain as other negative samples like $M_C$. 
Faulty negatives strongly repel the anchor and the negative sample, even though they should preferably be close in representation domain. \cite{robinson2021contrastive, huynh2020boosting}. 

To mitigate the effect of faulty negatives in CL, $M_A$ and $M_B$ should be ``less negative" comparing to $M_B$ and $M_C$ as illustrated in Figure~\ref{fig:overview}d. Namely, the latent vector of $M_A$ is not pushed too far away from $M_B$, while the agreement of $M_B$ and $M_C$ is still minimized during training. In particular, we propose a weighted NT-Xent loss $\mathcal{L}_{i,j}^{\text{w}}$ which penalizes each negative instance against the anchor via molecular similarities. The similarity measurement between two latent vectors $(\pmb z_i,\pmb z_k)$ from a negative molecule pair $(M_i, M_k)$ is penalized by a weight coefficient $w_{ik}$, as given in Equation~\ref{eq:weighted_ntxent}:
\begin{equation}
\begin{aligned}
\label{eq:weighted_ntxent}
    &\mathcal{L}_{i,j}^{\text{w}} = - \log \frac{\exp(\cos(\pmb{z}_i, \pmb z_j) / \tau)}{\sum_{k=1}^{2N} \mathbbm{1}_{k \neq i} \exp(w_{ik} \cos(\pmb{z}_i, \pmb z_k) / \tau)},
\end{aligned}
\end{equation}
where $w_{ik} \in [0,1]$. To identify the faulty negative instances, cheminformatic fingerprint is leveraged to evaluate the similarity between molecule pairs, as shown in Equation~\ref{eq:coeff}: 
\begin{equation}
    w_{ik} = 1 -\lambda_1 \texttt{FPSim}(M_i, M_k),
\label{eq:coeff}
\end{equation}
where $\texttt{FPSim}(M_i, M_k)$ evaluates the fingerprint similarity of the given two molecules $(M_i, M_k)$ and $\lambda_1$ is the hyperparameter that determines the scale of penalty for faulty negatives. In this work, we model $\texttt{FPSim}(\cdot,\cdot)$ as the Tanimoto similarity \cite{chen2002performance} of extended-connectivity fingerprint (ECFP) \cite{rogers2010extended}, since it has been demonstrated to be efficient and effective measurement of molecule similarities in multiple domains \cite{bajusz2015tanimoto}. Through the weighted NT-Xent loss, faulty negative molecule pairs, i.e., those with high fingerprint similarities, are forced to be closer in representation domain, this greatly mitigates faulty negatives and benefits the prediction of molecular properties. 

\subsection{Fragment Contrast}
\label{sec:fragment}

Most CL frameworks for molecular representation learning perform contrastive training on the whole molecule graph level \cite{wang2021molclr, zhu2021dual, liu2021pre, stark20213d}. Few works have investigated motif-level CL for molecules, which learns a table of frequently-occurring motif embeddings and trains a sampler to generate informative subgraphs for CL \cite{zhang2020motif}. Though such a method shows performance enhancement on various benchmarks, the learned sampler may not cover all the unique substructures in the large molecule dataset. Additionally, the previous work performs contrast across molecule graphs and motif subgraphs, which can lead to the ignorance of unsampled substructures while each molecule relies on a combination of different motifs to function. 
Driven by the insight, we leverage a widely used systematic fragmentation method, BRICS decomposition \cite{degen2008art}, to break each molecule into fragments, which preserves major structural features of chemical compounds \cite{liu2017break}, and perform CL on substructures individually from molecule-level contrast. Figure~\ref{fig:overview}e illustrates the proposed fragment contrast, where different colors indicate fragments obtained from BRICS. Instead of pooling over the entire graph, we conduct pooling on the decomposed subgraphs to obtain fragment representations. Different fragments, either from the same molecule or different molecules, are treated as negative pairs, like chlorobenzene (colored by purple) and benzofuran (colored by dark yellow). Notably, some fragments may share similar structures, however, they are still considered as negative instances. This is because different fragments have different neighbors to aggregate from. By this means, representations of fragments are forced to embed their unique neighboring information through layer-wise aggregation in GNNs. Such neighboring information can benefit molecular property prediction as molecules function differently due to the categorical and positional combination of various functional groups. 

Assume $2M$ fragments are created given a batch of $2N$ augmented molecule graphs. Notice that when conducting augmentations, fragments within molecules are also randomly transformed (i.e., atom masking and bond deletion). Thus, no extra augmentation is required to generate positive fragment pairs. A positive fragment pair $(F_i, F_j)$ is mapped to latent vectors $(\pmb z^{\text{frag}}_i, \pmb z^{\text{frag}}_j)$ through the same GNN encoder $f(\cdot)$ except the readout is conducted on the fragment subgraph instead of the whole molecule graph. The contrastive loss on the fragment level is given in Equation~\ref{eq:frag_ntxent}: 
\begin{equation}
    \mathcal{L}_{i,j}^{\text{frag}} = - \log \frac{\exp(\cos(\pmb{z}_i^{\text{frag}}, \pmb z_j^{\text{frag}}) / \tau)}{\sum_{k=1}^{2M} \mathbbm{1}_{k \neq i} \exp(\cos(\pmb{z}_i^{\text{frag}}, \pmb z_k^{\text{frag}}) / \tau)}.
\label{eq:frag_ntxent}
\end{equation}
Eventually, the total loss given in Equation~\ref{eq:total_loss} is a combination of the weighted contrastive loss on the whole molecule graph level shown in Equation~\ref{eq:weighted_ntxent}, and fragment level constrastive loss shown in \ref{eq:frag_ntxent} : 
\begin{equation}
    \mathcal{L}_{i,j}^{total} = \mathcal{L}_{i,j}^{\text{w}} + \lambda_2 \mathcal{L}_{i,j}^{\text{frag}},
\label{eq:total_loss}
\end{equation}
where $\lambda_2 \in (0,1]$ is a hyperparameter that controls the scale of fragment-level contrast during pre-training. 

\subsection{Datasets}
\label{sec:datasets}

The model is pre-trained on approximately 10 million unique unlabeled molecules from PubChem \cite{kim2019pubchem} collected and cleaned by Chithrananda et al \cite{chithrananda2020chemberta}. The pre-training dataset is randomly split into training and validation sets by the ratio of 95/5. The GNN model is pre-trained on the training set and tested on the validation set to select the best-performing model. 

To evaluate the performance of iMolCLR framework, we fine-tune the pre-trained GNN model on 12 benchmarks from MoleculeNet \cite{wu2018moleculenet}, including 7 classification and 5 regression benchmarks. These benchmarks contain a wide variety of molecular properties covering physiology, biophysics, physical chemistry, and quantum mechanics. During fine-tuning, each dataset is split into train/validation/test sets through scaffold split by the ratio of 80/10/10 following previous molecule SSL works \cite{Hu2020Strategies, rong2020self, wang2021molclr, zhu2021dual}. Comparing to random split, scaffold split provides a more challenging yet more realistic setting to benchmark molecular property predictions \cite{wu2018moleculenet}. During fine-tuning, the model is only trained on the train set and leverages the validation set to select the best-performing model. The performance of the selected model on the test set is reported in this work. More details of molecular property benchmarks can be found in Supplementary Table 1.

\subsection{Training Details}
\label{sec:training}

In iMolCLR pre-training, the GNN encoder embeds each molecule graph into a 512-dimension representation $\pmb h$. The projection head is modeled by an MLP with one hidden layer maps $\pmb h$ into 256-dimensional latent vector $\pmb z$. ReLU \cite{maas2013rectifier} is implemented as the non-linear activation function. The whole model is pre-trained for 50 epochs with batch size 512. We use Adam optimizer \cite{kingma2014adam} with an initial learning rate $5\times10^{-4}$ and the weight decay $1\times10^{-5}$. Additionally. cosine learning rate decay \cite{loshchilov2016sgdr} is performed during pre-training. 

During fine-tuning, we replace the projection head with a randomly initialized MLP which maps the representation $\pmb h$ into the desired property prediction while keeping the pre-trained GNN encoder. The pre-trained model is trained individually for 100 epochs on each task from the benchmarks. We perform a random search of hyperparameters on validation sets and report the results on test sets. For each benchmark, we run three individual runs and report the average and standard deviation of three trials. The whole model is implemented on PyTorch Geometric \cite{Fey/Lenssen/2019}. More details of fine-tuning hyperparameters can be found in Supplementary Table 2.

\subsection{Baselines}
\label{sec:baselines}

To demonstrate the effectiveness of iMolCLR, we compare its performance with various supervised GNN models. 
GCN \cite{kipf2016semi} and GIN \cite{xu2018how, Hu2020Strategies}, as prevalent GNN models for general graphical tasks, are implemented for performance comparison. Additionally, GNN models that are designed for molecular property prediction and have achieved state-of-the-art (SOTA) performance on certain benchmarks are included. D-MPNN \cite{yang2019analyzing} leverages a message-passing architecture that is invariant to molecule graph. SchNet \cite{schutt2018schnet} and MGCN \cite{lu2019molecular} model the quantum interactions within molecules in the graph aggregation. Additionally, attention-based model, AttentiveFP \cite{xiong2019pushing}, is included in baselines. 

We further compare our proposed method with other pre-training and SSL models. N-gram \cite{Liu2019NGramGS} obtains molecular representation through assembling the vertex embedding in short walks. Hu et al. \cite{Hu2020Strategies} is included, which contains both a self-supervised node-level and a supervised graph-level pre-training. MolCLR \cite{wang2021molclr} proposes a general CL framework for molecular representation learning. Notably, Hu et al., MolCLR, and our proposed iMolCLR are all implemented based on the GIN encoder. Thus, a comparison of these models, as well as supervised-learning GIN, well reflects the effectiveness of different SSL methods on various molecular property predictions.

%%%%%%%%%%%%%%%%%%%%%%%%%%%%%%%%%%%%%%%%%%%%%%%%%%%%%%%%%%%%%%%%%%%%%

\section{Results and Discussion}

\subsection{Molecular Property Predictions}

Molecular SSL methods are commonly evaluated by their predictive performance on various molecular properties. It is expected that good molecular representations learned through SSL greatly boost the prediction performance \cite{Hu2020Strategies, rong2020self, wang2021molclr}. Table~\ref{tb:classification} demonstrate the test compute area under the receiver operating characteristic curve (ROC-AUC) on classification benchmarks of our CL pre-training model in comparison to a wide variety of supervised (the first 6 models) and self-supervised (the last 4 models) baselines. Both best performing supervised and self-supervised models for each benchmark are highlighted in bold. The last column lists the averaged performance over all the classification benchmarks for each model. As shown in Table~\ref{tb:classification}, iMolCLR pre-training significantly boosts the performance of the GIN model by 14.2\% comparing to mere supervised learning. Built upon GIN, a generic GNN architecture, iMolCLR rivals strong supervised learning models and even exceeds them on 4 out of 7 classification benchmarks, where the latter develop sophisticated graph convolutional operations or engineered descriptors. For example, on SIDER and MUV, iMolCLR prevails over the best-performing supervised models by 6.7\% and 13.2\%, respectively. Furthermore, iMolCLR  demonstrates an overall preferable performance of 83.1\% ROC-AUC on average for all the datasets that we considered. Particularly, in comparison to MolCLR, the original CL framework with neither faulty negative mitigation nor fragment contrast, iMolCLR outperforms by 1.3\% ROC-AUC which demonstrates the effectiveness of improvement for CL in these challenging benchmarks. 

\begin{table}[t!]
  \centering
%   \small
    \footnotesize
  \begin{tabular}{l|lllllll|l}
    \toprule
    Dataset & BBBP & Tox21 & ClinTox & HIV & BACE & SIDER & MUV & Avg. \\
    \midrule
    GCN \cite{kipf2016semi} & 71.8$\pm$0.9 & 70.9$\pm$2.6 & 62.5$\pm$2.8 & 74.0$\pm$3.0 & 71.6$\pm$2.0 & 53.6$\pm$3.2 & 71.6$\pm$4.0 & 68.0 \\
    GIN \cite{xu2018how} & 65.8$\pm$4.5 & 74.0$\pm$0.8 & 68.2$\pm$3.7 & 75.3$\pm$1.9 & 70.1$\pm$5.4 & 57.3$\pm$1.6 & 71.8$\pm$2.5 & 68.9 \\
    SchNet \cite{schutt2018schnet} & 84.8$\pm$2.2 & 77.2$\pm$2.3 & 71.5$\pm$3.7 & 70.2$\pm$3.4 & 76.6$\pm$1.1 & 53.9$\pm$3.7 & 71.3$\pm$3.0 & 72.2 \\
    MGCN \cite{lu2019molecular} & 85.0$\pm$6.4 & 70.7$\pm$1.6 & 63.4$\pm$4.2 & 73.8$\pm$1.6  & 73.4$\pm$3.0 & 55.2$\pm$1.8 & 70.2$\pm$3.4 & 70.2 \\
    D-MPNN \cite{yang2019analyzing} & 81.2$\pm$3.8 & 78.9$\pm$1.3 & 90.5$\pm$5.3 & 75.0$\pm$2.1 & 85.3$\pm$5.3 & \bftab{63.2$\pm$2.3} & 76.2$\pm$2.8 & 78.7 \\
    AttentiveFP \cite{xiong2019pushing} & \bftab{90.8$\pm$5.0} & \bftab{80.7$\pm$2.0} & \bftab{93.3$\pm$2.0} & \bftab{82.9$\pm$2.2} & \bftab{86.3$\pm$1.5} & 60.5$\pm$6.0 & \bftab{77.6$\pm$3.1} & \bftab{81.7} \\
    \midrule
    N-Gram \cite{Liu2019NGramGS} & \bftab{91.2$\pm$3.0} & 76.9$\pm$2.7 & 85.5$\pm$3.7 & \bftab{83.0$\pm$1.3} & 87.6$\pm$3.5 & 63.2$\pm$0.5 & 81.6$\pm$1.9 & 81.3 \\
    Hu et al. \cite{Hu2020Strategies} & 70.8$\pm$1.5 & 78.7$\pm$0.4 & 78.9$\pm$2.4 & 80.2$\pm$0.9 & 85.9$\pm$0.8 & 65.2$\pm$0.9 & 81.4$\pm$2.0 & 77.3 \\
    MolCLR \cite{wang2021molclr} & 73.6$\pm$0.5 & \bftab{79.8$\pm$0.7} & 93.2$\pm$1.7 & 80.6$\pm$1.1 & \bftab{89.0$\pm$0.3} & 68.0$\pm$1.1 & 88.6$\pm$2.2 & 81.8 \\
    iMolCLR & 76.4$\pm$0.7 & \bftab{79.9$\pm$0.6} & \bftab{95.4$\pm$1.1} & 80.8$\pm$0.1 & 88.5$\pm$0.5 & \bftab{69.9$\pm$1.5} & \bftab{90.8$\pm$1.7} & \bftab{83.1} \\
    \bottomrule
  \end{tabular}
  \caption{Mean and standard deviation of test ROC-AUC (\%) of iMolCLR in comparison to different supervised and self-supervised learning models on classification benchmarks.}
  \label{tb:classification}
\end{table}

Additionally, we test the performance of our CL pre-training model and baselines on 5 regression benchmarks as demonstrated in Table~\ref{tb:regression}. The last column of the table shows the average of scaled error over all the regression benchmarks for each model, which is calculated by dividing the error by the range of each database label. We report root mean square error (RMSE) for FreeSolv, ESOL, and Lipo and mean absolute error (MAE) for QM7 and QM8 following MoleculeNet \cite{wu2018moleculenet}. Similar to the performance on classification benchmarks, iMolCLR also demonstrates a rival or even superior prediction accuracy over supervised baseline models on challenging regression benchmarks. In terms of the averaged scaled error, iMolCLR outperforms the best-supervised model, D-MPNN, by 0.0019, which is a nontrivial improvement as regression tasks are more challenging than classifications. In comparison with the original CL pre-training, iMolCLR decreases the prediction errors on 4 out of 5 benchmarks and shows competitive performance on the remaining ESOL dataset. Other SSL baselines also cannot emulate iMolCLR on most datasets. In particular, iMolCLR shows an advantage of 0.0157 and 0.0239 on scaled error over N-Gram \cite{Liu2019NGramGS} and Hu et al. \cite{Hu2020Strategies}, respectively. 

\begin{table}[t!]
  \centering
  \small
%   \footnotesize
  \begin{tabular}{l|lllll|l}
    \toprule
    Dataset & FreeSolv & ESOL & Lipo & QM7 & QM8 & Scaled avg.\\
    \midrule
    GCN \cite{kipf2016semi} & 2.87$\pm$0.14 & 1.43$\pm$0.05 & 0.85$\pm$0.08 & 122.9$\pm$2.2 & 0.0366$\pm$0.0011 & 0.1002 \\
    GIN \cite{xu2018how} & 2.76$\pm$0.18 & 1.45$\pm$0.02 & 0.85$\pm$0.07 & 124.8$\pm$0.7 & 0.0371$\pm$0.0009 & 0.1002 \\
    SchNet \cite{schutt2018schnet} & 3.22$\pm$0.76 & 1.05$\pm$0.06 & 0.91$\pm$0.10 & \bftab{74.2$\pm$6.0} & 0.0204$\pm$0.0021 & 0.0861 \\
    MGCN \cite{lu2019molecular} & 3.35$\pm$0.01 & 1.27$\pm$0.15 & 1.11$\pm$0.04 & 77.6$\pm$4.7 & 0.0223$\pm$0.0021 & 0.0982 \\
    D-MPNN \cite{yang2019analyzing} & 2.18$\pm$0.91 & 0.98$\pm$0.26 & \bftab{0.65$\pm$0.05} & 105.8$\pm$13.2 & \bftab{0.0143$\pm$0.0022} & \bftab{0.0699} \\
    AttentiveFP \cite{xiong2019pushing} & \bftab{2.03$\pm$0.42} & \bftab{0.85$\pm$0.06} & \bftab{0.65$\pm$0.03} & 126.7$\pm$4.0 & 0.0282$\pm$0.0010 & 0.0755 \\
    \midrule
    N-Gram \cite{Liu2019NGramGS} & 2.51$\pm$0.19 & \bftab{1.10$\pm$0.03} & 0.88$\pm$0.12 & 125.6$\pm$1.5 & 0.0320$\pm$0.0032 & 0.0919 \\
    Hu et al. \cite{Hu2020Strategies} & 2.83$\pm$0.12 & 1.22$\pm$0.02 & 0.74$\pm$0.00 & 110.2$\pm$6.4 & 0.0191$\pm$0.0003 & 0.0837 \\
    MolCLR \cite{wang2021molclr} & 2.20$\pm$0.20 & \bftab{1.11$\pm$0.01} & \bftab{0.65$\pm$0.08} & 87.2$\pm$2.0 & 0.0174$\pm$0.0013 & 0.0714\\
    iMolCLR & \bftab{2.09$\pm$0.03} & 1.13$\pm$0.02 & \bftab{0.64$\pm$0.00} &	\bftab{66.3$\pm$2.0} & \bftab{0.0170$\pm$0.0002} & \bftab{0.0680} \\
    \bottomrule
  \end{tabular}
  \caption{Mean and standard deviation of test RMSE (for FreeSolv, ESOL, Lipo) or MAE (for QM7, QM8) of iMolCLR in comparison to different supervised and self-supervised learning models on regression benchmarks. The scaled error is calculated by dividing RMSE/MAE by the range of each benchmark labels. }
  \label{tb:regression}
\end{table}

Overall, experiments on a wide variety of challenging molecular property benchmarks demonstrate that our proposed iMolCLR is an effective SSL strategy that greatly improves the performance of pre-trained GNN models. Further, iMolCLR shows superiority over the original CL framework on both classification and regression tasks, which validates the effectiveness of faulty negative mitigation and fragment contrast. The following section investigates how the two techniques impact the molecular CL methods in detail. 

\subsection{Influence of Faulty Negative Mitigation and Fragment Contrast in Molecular CL}

\begin{figure}[t!]
    \centering
    \includegraphics[width=\textwidth, keepaspectratio=true]{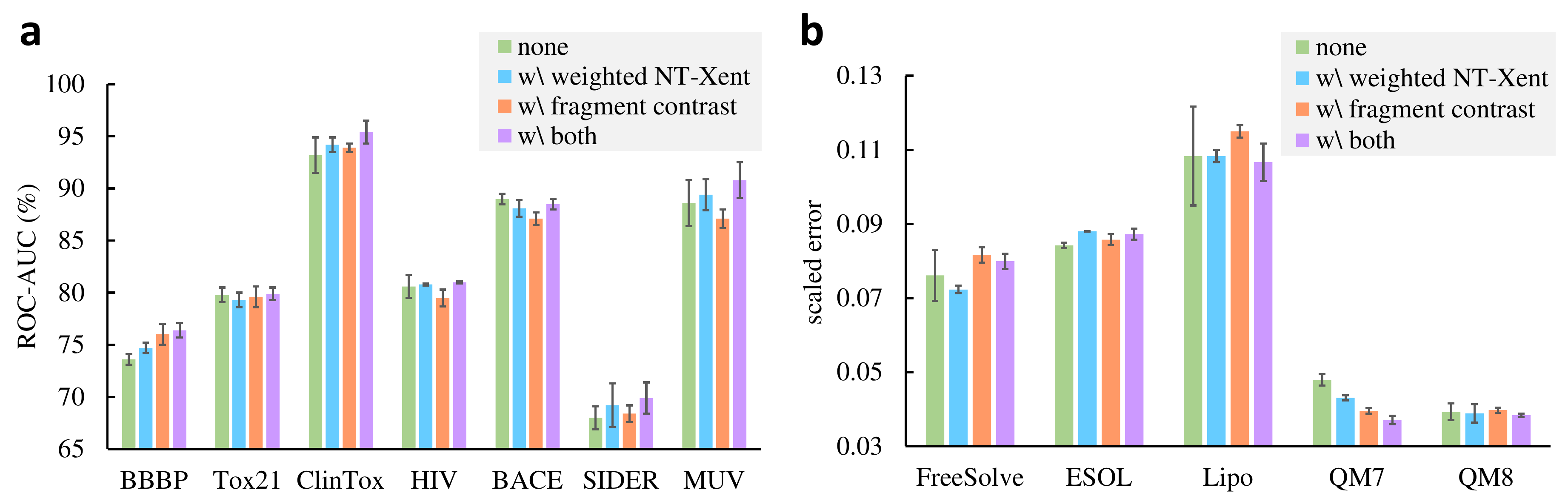}
    \caption{Investigation of the impact of weighted NT-Xent and fragment contrast on molecular CL pre-training. (a) Mean and standard deviation of test ROC-AUC ($\uparrow$) for different CL strategies on classification benchmarks. (b) Mean and standard deviation of scaled error ($\downarrow$) for different CL strategies on regression benchmarks.}
    \label{fig:ablation}
\end{figure}

We further probe the impact of the two improvement strategies: faulty negative mitigation and fragment contrast on molecular CL. Specifically, whether contrastive training benefits from each method solely or the combination of both strategies. To this end, we consider four pre-training strategies: original CL pre-training, CL with only weighted NT-Xent for faulty negative mitigation, CL with only fragment contrast, and CL with both improvement methods, as shown in Figure~\ref{fig:ablation}. The four strategies are illustrated by green, blue, orange, and purple bars, respectively. Figure~\ref{fig:ablation}a shows ROC-AUC on classification tasks and Figure~\ref{fig:ablation}b shows scaled error on regression tasks. The height of each bar denotes the averaged performance while the length of each error bar represents the standard deviation over three individual runs. It is illustrated that the integration of weighted NT-Xent and fragment contrast demonstrates the best performance on 6 out 7 classification benchmarks and 3 out of 5 regression benchmarks.
On average, iMolCLR with both improvement strategies applied surpasses weighted NT-Xent and fragment contrast solely by 0.9\% and 1.5\% on classifications, respectively. Similarly, iMolCLR shows a decrease of the averaged scaled error by 0.0003 and 0.0025 over each strategy alone on regressions. Also, weighted NT-Xent solely improves the property prediction over original CL on most benchmarks except for Tox21, BACE, and ESOL. Interestingly, fragment contrast alone shows a limited advantage over original CL pre-training. This could be because when applying only fragment contrast, decomposed substructures are considered as negative pairs, this may exacerbate the faulty negatives in contrastive training. On the other hand, with faulty negative mitigated through weighted NT-Xent loss, through fragment contrast the backbone model can easily identify the functional groups or motifs within each molecule. Thus, the weighted NT-Xent mitigates the faulty negative instances not only on the molecule level, but also on the decomposed fragment level. Overall, the combination of faulty negative mitigation together with fragment contrast greatly improves the molecular CL for property prediction and it demonstrates an advantage over applying each strategy solely. More test results of different pre-training strategies can be found in Supplementary Tables 3 and 4. 

\subsection{Empirical Study of Hyperparameters}

To better investigate the proposed strategies for molecular CL, we conduct empirical study of hyperparameters of contrastive loss given in Equation~\ref{eq:weighted_ntxent}, \ref{eq:coeff}, \ref{eq:frag_ntxent}, and \ref{eq:total_loss}. Different combinations of $\lambda_1$, $\lambda_2$, and $\tau$ are tested on molecular property prediction benchmarks. $\lambda_1$ controls the scale of penalty on faulty negative instances and $\lambda_2$ weighs the magnitude of fragment-level contrast. Besides, the selection of the appropriate $\tau$ benefits learning from hard negative samples \cite{chen2020simple}. Table~\ref{tb:hyperparam} shows the performance of iMolCLR pre-trained GNNs with different combinations of the three hyperparameters on both classification and regression benchmarks. $\lambda_1$ and $\lambda_2$ are selected from $\{0.3, 0.5, 0.7\}$, and $\tau$ is selected from $\{0.05, 0.1, 0.5\}$. iMolCLR achieves the best overall performance on classifications under $\lambda_1=0.5, \lambda_2=0.5, \tau=0.1$, while on regressions, using $\lambda_1=0.5, \lambda_2=0.5, \tau=0.5$ obtains the best results. Tuning each hyperparameter may have an opposite impact on different benchmarks. For instance, decreasing $\lambda_1$ from 0.5 to 0.3 causes a drop of 0.8\% ROC-AUC on classifications, while a gain of 0.0008 is observed on regressions. It indicates that the best combination of hyperparameters is task-dependent on the target property and data distribution. However, it should be pointed out that though the selection of hyperparameters affects the performance on downstream molecular property predictions as demonstrated in Table~\ref{tb:hyperparam}, all the listed hyperparameter combinations still significantly boost GNNs in comparison to supervised learning. This reflects the robustness of our proposed molecular CL framework in learning expressive representations. Detailed test results of property prediction on each benchmark can be found in Supplementary Tables 3 and 4. 

\begin{table}[t!]
\centering
\small
\caption{Evaluation for iMolCLR pre-trained GNNs with different combinations of $\lambda_1$, $\lambda_2$, and $\tau$. Both averaged ROC-AUC over classification benchmarks and averaged error over regression benchmarks are reported.}
\label{tb:hyperparam}
    \begin{tabular}{c|ccccccc}
    \toprule
    $\lambda_1$ & 0.5 & 0.3 & 0.7 & 0.5 & 0.5 & 0.5 & 0.5 \\
    $\lambda_2$ & 0.5 & 0.5 & 0.5 & 0.3 & 0.7 & 0.5 & 0.5 \\
    $\tau$ & 0.1 & 0.1 & 0.1 & 0.1 & 0.1 & 0.05 & 0.5\\
    \midrule
    Avg. ROC-AUC (\%) ($\uparrow$) & \bftab{82.8} & 82.0 & 81.6 & 82.3 & 82.0 & 81.0 & 81.4 \\
    Avg. scaled error ($\downarrow$) & 0.0733 & 0.0725 & 0.0730 & 0.0726 & 0.0723 & 0.0736 & \bftab{0.0712} \\
    \bottomrule
    \end{tabular}
\end{table}

\subsection{Does Extra Features Benefit Molecular CL Pre-training?}

We implement simple yet distinguishable node and edge features via RDKit \cite{greg2006rdkit} to model 2D molecule graphs following previous pre-training frameworks \cite{Hu2020Strategies, wang2021molclr}. In particular, node features include atomic number and chirality type, and edge features consist of covalent bond type and direction. However, extra features can also be considered in molecule graphs \cite{xiong2019pushing}. In supervised learning, rich input features are expected to benefit molecular property predictions as more information is provided. This leads to the question: whether enriched input benefits molecular CL pre-training for property predictions? To this end, we introduce more node and edge features to iMolCLR pre-training as shown in Table~\ref{tb:extra_features}. Besides the original features, degree, charge, hybridization, aromatic, and number of hydrogens are included in the node feature, meanwhile, the stereotype of bond is added to edge features. The extra features are fed into embedding layers and added to the embedding from original features before being sent to graph convolutional layers. During augmentation, each feature of masked nodes is set to a unique code. For example, \texttt{atomic} is set to 0 when masked. Through adding richer features, more molecular information is provided during pre-training as well as downstream fine-tuning. 

\begin{table}[t!]
  \centering
  \small
  \begin{tabular}{llll}
    \toprule
    Type & Name & Description & Range \\
    \midrule
    \multirow{7}{*}{Node} & \texttt{atomic} & Atomic number & $\{x:1 \leq x \leq 119, x \in \mathbbm{Z}\}$ \\
    & \texttt{chirality} & Chirality type & \{unspecified, CW, CCW, other\} \\
    & \texttt{degree} & Number of bonded neighbors & $\{x:0 \leq x \leq 10, x \in \mathbbm{Z}\}$ \\
    & \texttt{charge} & Formal charge of the atom & $\{x:-5 \leq x \leq 5, x \in \mathbbm{Z}\}$ \\
    & \texttt{hybrization} & Hybrization type & \{sp, sp\textsuperscript{2}, sp\textsuperscript{3}, sp\textsuperscript{3}d, sp\textsuperscript{3}d\textsuperscript{2}, other\} \\
    & \texttt{aromatic} & Whether on a aromatic ring & $\{0,1\}$ \\
    & \texttt{hydrogen} & Number of bonded hydrogens & $\{x:0 \leq x \leq 5, x \in \mathbbm{Z}\}$ \\
    \midrule
    \multirow{3}{*}{Edge} & \texttt{bond\_type} & Type of covalent bonds & \{single, double, triple, aromatic\} \\
    & \texttt{bond\_dir} & Direction of covalent bonds & \{none, end-upright, end-downright\} \\
    & \texttt{stereo} & Stereotype & \{None, Any, Z, E, Cis, Trans\}  \\
    \bottomrule
  \end{tabular}
  \caption{List of extended node and edge features for molecule graphs.}
  \label{tb:extra_features}
\end{table}

\begin{figure}[t!]
    \centering
    \includegraphics[width=\textwidth, keepaspectratio=true]{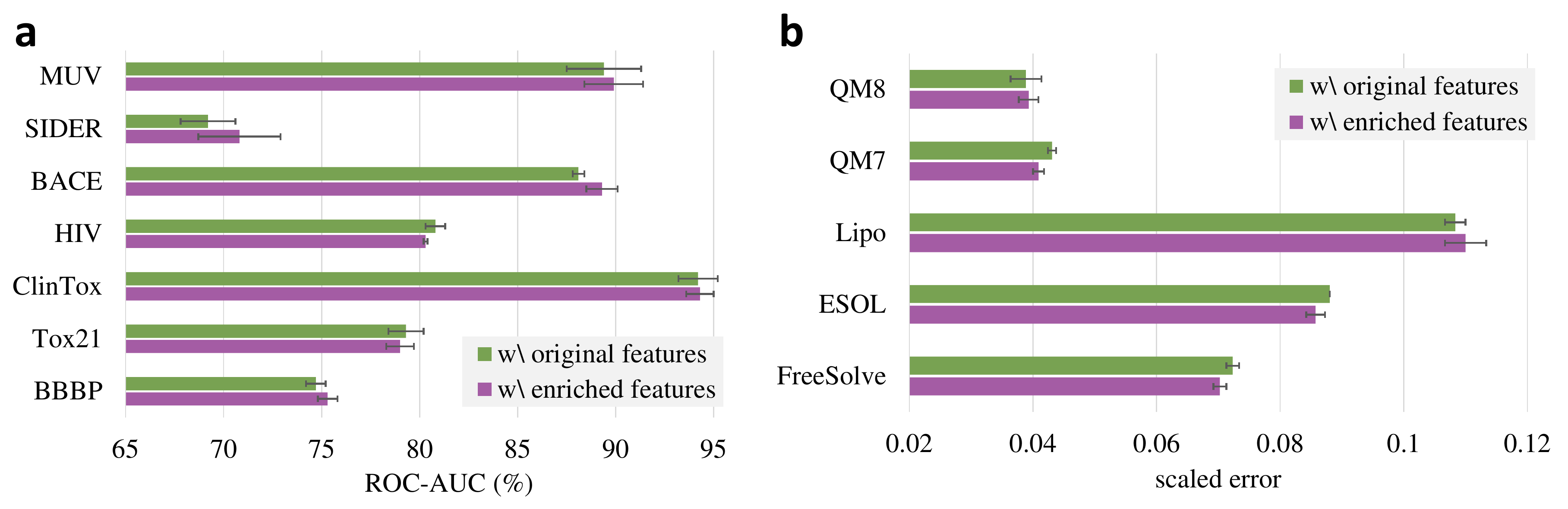}
    \caption{Comparison of molecular CL pre-training with original and enriched input features. (a) Test ROC-AUC ($\uparrow$) of pre-trained GNNs on classification benchmarks. (b) Test scaled error ($\downarrow$) of pre-trained GNNs on regression benchmarks.}
    \label{fig:enrich_feat}
\end{figure}

We compare the results of GNN models pre-trained and fine-tuned with original and enriched features. As shown in Figure~\ref{fig:enrich_feat}, we implement molecular CL pre-training with weighted NT-Xent loss, where green bars represent training using original input features while purple bars represent enriched features. Both the mean and standard deviation of performance on each benchmark are illustrated in Figure~\ref{fig:enrich_feat}. On most benchmarks, adding extra input features benefits the downstream molecular property prediction. For instance, enriched features improve ROC-AUC by 1.6\% on SIDER and reduce MAE by 3.9 on QM7. Although adding extra features demonstrates better performances on several benchmarks, the improvements are limited comparing to the various categories of extra features included. In few benchmarks, CL with enriched features even slightly falls behind the original model like on HIV and Tox21. This reveals that our molecular CL framework effectively learns the intrinsic relationships between atoms without heavily relying on engineered features. Extra node features, like \texttt{degree} (i.e., number of neighbors) and \texttt{hydrogen} (i.e., number of neighboring hydrogen atoms), can be inherently learned by pre-trained graph aggregations without explicit provided. Other features such as \texttt{hybrization} and \texttt{charge} are strongly related with the \texttt{atomic} feature, providing a redundant input for CL. The major takeaway is that our proposed molecular CL, as a self-supervised pre-training strategy, learns expressive representations from graphs even, which does not require engineered and enriched features to achieve better performance. 

\subsection{Case Study of iMolCLR Representations}

\begin{figure}[t!]
    \centering
    \includegraphics[width=\textwidth, keepaspectratio=true]{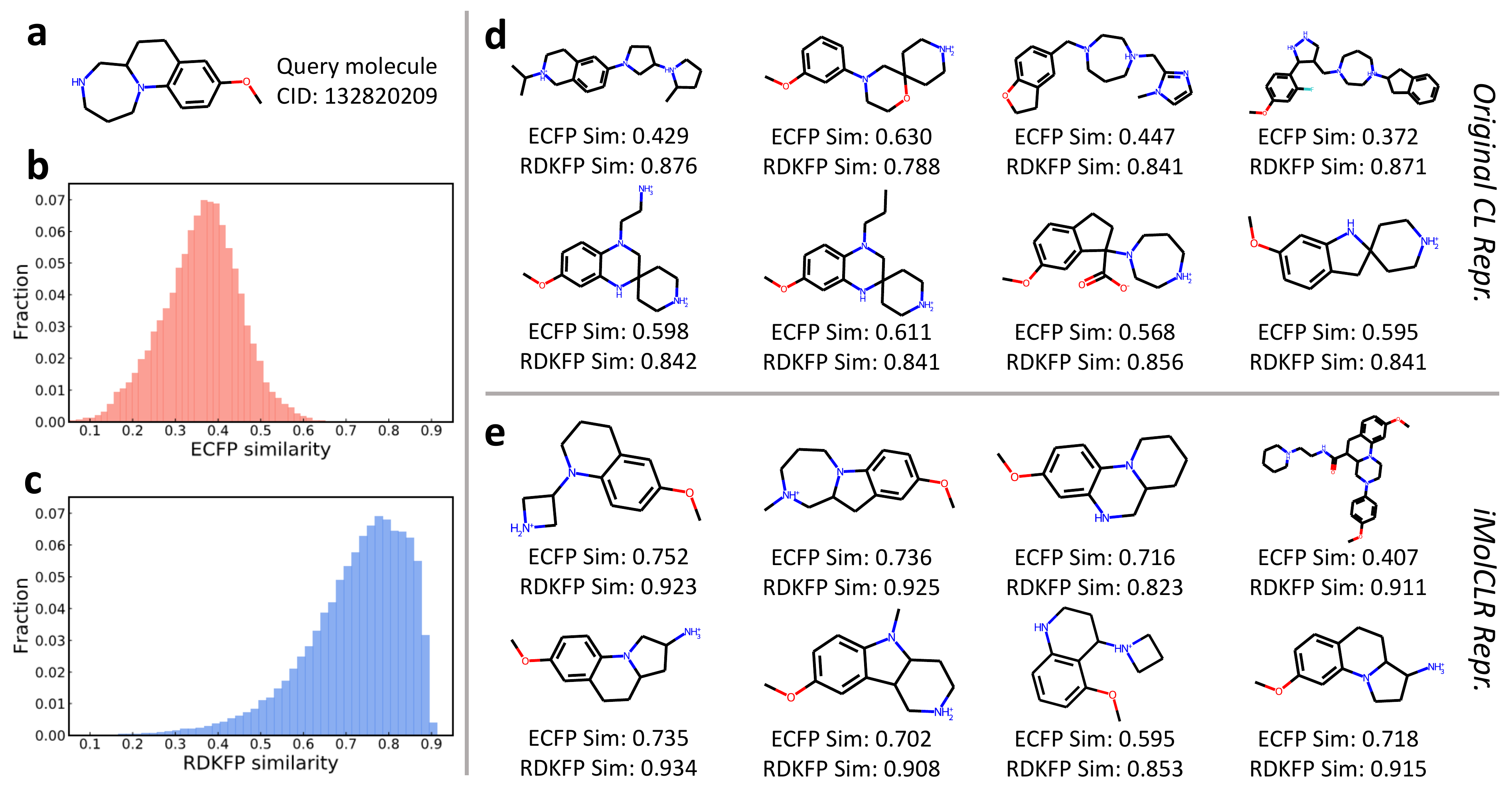}
    \caption{Case study of CL learned molecular representations. (a) Query molecule of CID 132820209. (b) Distribution of Tanimoto similarities on ECFP between molecules from the pre-training dataset and the query. (c) Distribution of Tanimoto similarities on RDKFP between molecules from the pre-training dataset and the query. (d) 8 molecules that are closest to the query on the original CL learned representation domain with Tanimoto similarities on ECFP and RDKFP denoted. (e) 8 molecules that are closest to the query on the iMolCLR learned representation domain with Tanimoto similarities on ECFP and RDKFP denoted.}
    \label{fig:mol_retrival}
\end{figure}

To further evaluate iMolCLR, we compare the molecular representations learned by iMolCLR with those learned by original CL together with cheminformatics fingerprints. Given the query molecule (CID: 132820209) shown in Figure~\ref{fig:mol_retrival}a, we compute the Tanimoto similarities \cite{chen2002performance} of cheminformatics fingerprints between the query and all the molecules in the $\sim$10M pre-training dataset. Figure~\ref{fig:mol_retrival}b and \ref{fig:mol_retrival}c exhibit the distribution of similarities on extended-connectivity fingerprint (ECFP) \cite{rogers2010extended} and RDKit-specific fingerprint (RDKFP) \cite{greg2006rdkit}, respectively. ECFP is a topological fingerprint for structure−activity modeling while RDKFP identifies subgraphs of different sizes. Overall, ECFP leads to lower similarity scores than RDKFP due to the different features and algorithms implemented. Within the database, molecules are considered very similar to the query with respect to ECFP if their similarity is greater than 0.6 (Figure~\ref{fig:mol_retrival}b). While with regard to RDKFP, similar molecules are expected to have similarities greater than 0.9 (Figure~\ref{fig:mol_retrival}c).
We then select 8 molecules that are closest to the query in the representation domain learned by either original CL (Figure~\ref{fig:mol_retrival}d) or iMolCLR (Figure~\ref{fig:mol_retrival}e). Cosine similarity is used to measure the distances between learned representations. The 8 molecules selected by original CL have averaged ECFP similarity of 0.540 and RDKFP of 0.839, while those selected by iMolCLR have remarkably higher averaged ECFP similarity of 0.655 and RDKFP of 0.893. It is indicated that through faulty negative mitigation with weighted NT-Xent, iMolCLR embeds molecular cheminformatics in the learned representations, which can reason molecular similarities. Additionally, benefiting from the fragment contrast, substructure-level topology is also embedded in iMolCLR representations. For instance, almost all the closest molecules found by iMolCLR share the phenanthroline-like substructure of three fused rings with the query molecule, whereas the original CL retrieves molecules with the substructure of only two rings fused. Besides, the C-O-C substructure of the query is captured by iMolCLR and shared among all the selected molecules. Through the case study, iMolCLR demonstrates the improvement over original CL on the learned representations, which better embed cheminformatics and substructure topology. More examples of molecule retrieval through iMolCLR can be found in Supplementary Figures 1, 2, and 3.

%%%%%%%%%%%%%%%%%%%%%%%%%%%%%%%%%%%%%%%%%%%%%%%%%%%%%%%%%%%%%%%%%%%%%

\section{Conclusions}

In this work, we propose iMolCLR, an improvement of molecular contrastive learning of representations with GNNs. Specifically, two strategies are introduced in iMolCLR: (1) the weighted NT-Xent loss to mitigate faulty negative instances during contrastive pre-training, (2) fragment-level contrast on substructures from BRICS decomposition. The former considers cheminformatics such that learned representations are related to molecular similarities, which are neglected by previous molecular CL methods. The latter, on the other hand, encourages the pre-trained GNNs to embed functional groups and motif information which are vital to molecular properties. Benefiting from the two strategies, iMolCLR outperforms other SSL baseline models, including the original molecular CL, on a wide variety of molecular property prediction benchmarks. Further investigation demonstrates that iMolCLR is an effective and robust pre-training framework, which learns expressive representations from limited input features. iMolCLR, an SSL method that can leverage large unlabeled data, bears a promise for accurate molecular property prediction, which can greatly benefit applications like drug and material discovery. 

% \bibliography{reference}

\providecommand{\latin}[1]{#1}
\makeatletter
\providecommand{\doi}
  {\begingroup\let\do\@makeother\dospecials
  \catcode`\{=1 \catcode`\}=2 \doi@aux}
\providecommand{\doi@aux}[1]{\endgroup\texttt{#1}}
\makeatother
\providecommand*\mcitethebibliography{\thebibliography}
\csname @ifundefined\endcsname{endmcitethebibliography}
  {\let\endmcitethebibliography\endthebibliography}{}

\end{document}